\documentclass[a4paper,11pt]{article} 
\usepackage[left=3.5cm, right=3.5cm, top=3.5cm, bottom=3.5cm]{geometry}
\usepackage[colorlinks=True, citecolor=blue]{hyperref}

\usepackage[utf8]{inputenc}
\usepackage{graphicx}
\usepackage{amsfonts}
\usepackage{natbib}
\usepackage[lined,ruled,commentsnumbered]{algorithm2e}
\usepackage{tikz}
\usetikzlibrary{fit,positioning}
\usepackage{url}
\usepackage{stmaryrd}
\usepackage{xcolor}
\usepackage{amsmath}
\usepackage{amsthm}
\usepackage{bbm}
\usepackage{ dsfont }

\DeclareMathOperator*{\argmin}{argmin}

\newcommand{\W}{\mathbf{W}}
\newcommand{\Hbf}{\mathbf{H}}
\newcommand{\Y}{\mathbf{Y}}
\newcommand{\U}{\mathbf{U}}
\newcommand{\V}{\mathbf{V}}
\newcommand{\A}{\mathbf{A}}
\newcommand{\C}{\mathbf{C}}
\newcommand{\cbf}{\mathbf{c}}

\def\bal#1{\begin{align}#1\end{align}}
\newcommand{\ve}[1]{ {\mathbf{#1}} }
\renewcommand{\H}{\ve{H}}

\begin{document}

\title{Negative Binomial Matrix Factorization for Recommender Systems}

\author{ Olivier Gouvert \qquad  Thomas Oberlin \qquad Cédric Févotte \\
\small IRIT, Université de Toulouse, CNRS, France }

\date{}

\maketitle

\begin{abstract}
We introduce negative binomial matrix factorization~(NBMF), a matrix factorization technique specially designed for analyzing over-dispersed count data. It can be viewed as an extension of Poisson matrix factorization (PF) perturbed by a multiplicative term which models exposure. This term brings a degree of freedom for controlling the dispersion, making NBMF more robust to outliers. We show that NBMF allows to skip traditional pre-processing stages, such as binarization, which lead to loss of information. Two estimation approaches are presented: maximum likelihood and variational Bayes inference. We test our model with a recommendation task and show its ability to predict user tastes with better precision than PF. 
\end{abstract}

\section{Introduction}

Poisson matrix factorization~(PF) is a non-negative matrix factorization (NMF) model~\citep{lee_learning_1999} often used for recommender systems~\citep{ma_probabilistic_2011,gopalan_scalable_2015-1}, text information retrieval~\citep{canny_gap:_2004,buntine_discrete_2006} or dictionary learning for image processing~\citep{cemgil_bayesian_2009}. The data is assumed to be drawn from the Poisson distribution making it specially well suited for count/integer-valued data.

Since the Netflix Prize~\citep{bennett_netflix_2007}, collaborative filtering (CF) has been giving the state-of-the-art results for recommender systems. CF exploits data relating users to items, like historical data. These data can either be explicit (ratings given by users to items) or implicit (count data from users listening to songs, clicking on web pages, watching videos, etc). Count data can be summarized into a matrix $\Y\in\mathbb{N}^{U\times I}$, where $y_{ui}$ corresponds to the number of times a user $u$ interacts with an item $i$, $U$ and $I$ are the number of users and items respectively, and $\mathbb{N}$ is the set of integer values.

PF assumes that the observed data $\Y$ is generated from the process:
\[ y_{ui} \sim \operatorname{Pois}([\W\Hbf^T]_{ui}),\]
where $\W\in\mathbb{R}^{U\times K}_+$ represents the preferences of users and $\Hbf\in\mathbb{R}^{I\times K}_+$ represents the attributes of items~\citep{koren_matrix_2009}. Usually, $K\ll\min(U,I)$ which implies a low rank data approximation.

A limitation of using the Poisson distribution is that the variance is fixed and equal to the mean: $\operatorname{var}[y_{ui}] = \mathbb{E}[y_{ui}]$, making it poorly adapted for over-dispersed data. Yet, this happens to be the case with implicit data. As a matter of fact, this type of data, easy to collect, is known to be very sparse, noisy and bursty~\citep{hu_collaborative_2008,schein_poisson-gamma_2016}. As explained in~\citet{basbug_hierarchical_2016}, PF suffers from a strong coupling between the support of $\Y$ and its values. To remain robust to outliers, a pre-processing stage is often used~\citep{gopalan_scalable_2015-1,liang_modeling_2015}. Often, all positive values are thresholded to $1$, producing binary data, i.e., $\Y\in\{0,1\}^{U\times I}$.

We propose a new probabilistic matrix factorization (MF) model, coined negative binomial matrix factorization (NBMF), to avoid such a pre-processing stage that leads to information loss. NBMF offers a new degree of freedom for controlling data dispersion and can used with raw data. 

In Section~\ref{sec:nbmf}, we introduce NBMF and its connections with the state of the art. In Section~\ref{sec:mle}, we study the maximum likelihood estimator of NBMF and discuss the fit function/divergence it implies. In Section~\ref{sec:bayesian nbmf}, a Bayesian NBMF model for recommender systems and an inference mechanism are developed. Finally, we illustrate the benefits of NBMF with experiments on the music Taste Profile dataset \citep{bertin-mahieux_million_2011}.

\section{Negative Binomial Matrix Factorization} \label{sec:nbmf}

\subsection{Model}
We assume that, for each $u\in\{ 1,..,U\}$ and $i\in\{ 1,..,I\}$, our observations $y_{ui}=[\Y]_{ui}$ are sampled from the generative process
\begin{align}
y_{ui}\sim \operatorname{NB}\left(\alpha,\dfrac{1}{1+{\alpha}/{[\W\Hbf^T]_{ui}}}\right), \label{NB}
\end{align}
where $\operatorname{NB}(\alpha,p)$ is the negative binomial (NB) distribution parametrized by a dispersion coefficient $\alpha\in\mathbb{R}_+$ and a probability parameter $p\in [0,1]$. Its probability mass function is given by: 
\[\mathbb{P}(Y=y)={\frac {\Gamma (y+\alpha)}{y!\ \Gamma (\alpha)}}p^{y}(1-p)^{\alpha}.\] 

Like in Poisson factorization and many mean-parametrized matrix factorization models \citep{ardnmfj}, the expected value of the observations is given by: $\mathbb{E}[y_{ui}] = [\W\Hbf^T]_{ui}$, which gives an intuitive understanding of the model. Contrary to the Poisson distribution, the NB distribution has a second parameter $\alpha$ which enables to add variance to the model: 
\[\operatorname{var}[y_{ui}]= [\W\Hbf^T]_{ui}(1+\frac{[\W\Hbf^T]_{ui}}{\alpha})>\mathbb{E}[y_{ui}].\]

Note that PF is a particular case of our model, corresponding to the limit case $\alpha\to\infty$.

\begin{figure}[t]
\centerline{\includegraphics[height=6cm]{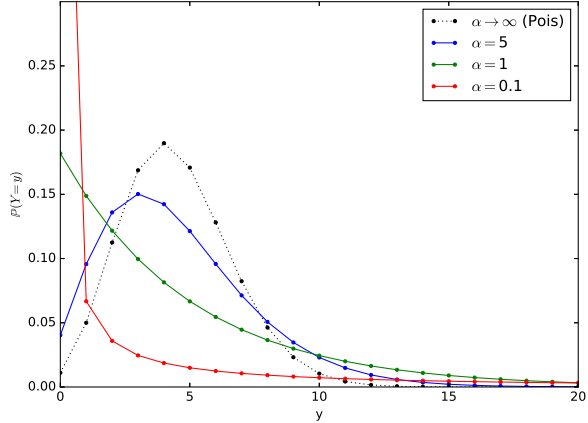}}
\caption{Influence of $\alpha$ on probability mass function of NB distribution. For $\alpha \leq 1$ the mode is $0$. When $\alpha$ goes to infinity, we recover the Poisson distribution.}
\end{figure}

The NB distribution can also be viewed as a Poisson-gamma mixture.\footnote{ We use the following convention for gamma distribution: $\mathcal{G}(x;\alpha,\beta)=x^{\alpha-1}e^{-\beta x}\beta^\alpha \Gamma(\alpha)^{-1}$, where $\alpha$ is the scale parameter and $\beta$ is the rate parameter.} Using this property, we can write the following equivalent hierarchical model:
\begin{align}
a_{ui}&\sim \mathcal{G}(\alpha,{\alpha}) \\ 
y_{ui}|a_{ui} &\sim \operatorname{Pois}(a_{ui}[\W\Hbf^T]_{ui}),
\end{align}
where the latent variables $a_{ui}$ control local variabilities. We denote by $\A$ the $U\times I$ matrix with coefficients $[\A]_{ui}=a_{ui}$. By construction, we have $\mathbb{E}[a_{ui}]=1$ and $\operatorname{var}(a_{ui})=\alpha^{-1}$.

\subsection{Interpretation of the Latent Variable $\A$} \label{sec:interpretation A}
The matrix $\A$ captures local variations that cannot be explained by the product $\W\Hbf^T$. $\A$ can attenuate or accentuate the entries of $\W\Hbf^T$. In the field of recommender systems, $\A$ can be viewed as an exposure variable.  In~\citet{liang_modeling_2015}, a similar variable was introduced. It modeled whether a user knows an item or not. This variable was sampled from a Bernoulli distribution, its values were therefore binary: $a_{ui}\in\{0,1\}$. Here, we have $a_{ui}\in\mathbb{R}^+$,  which allows for finest interpretations:
\begin{itemize}
\item If $a_{ui}\ll 1$, the user is under-exposed to the item. It may be explained by several reasons: the user does not frequent the places/communities where the song is played, he is not aware of the release of a new song, etc.
\item If $a_{ui}\gg 1$, the user is over-exposed to the item. For song recommendation, this over-exposure can be ``active'', e.g., the user listens to the song on repeat, or ``passive'', e.g., the item is heavily broadcasted on the radio, is highlighted on a website, etc.
\item If $a_{ui} \approx 1$, the exposure does not affect the listening pattern of the user which is fully described by $\W\Hbf^T$.
\end{itemize}

\subsection{Recommendation Task} \label{sec:reco task}

The goal of CF is to propose each user a personalized list of new items (items he has not consumed yet) that he may like. 

MF-based CF was first studied with explicit feedbacks \citep{koren_matrix_2009}. In explicit feedback, the observed matrix $\Y$ contains the ratings of users to items. These ratings express an explicit preference (positive or negative feedback). The goal of CF is therefore to predict, for each user, the feedbacks of unrated items (missing feedbacks). Recommendation is turned into a matrix completion problem.

This is not the case anymore when we work with implicit data (like count data). In fact, we can not differentiate missing and negative feedbacks. There is an ambiguity on the items which have never been consumed, i.e., the items such that $y_{ui}=0$. When $y_{ui}=0$, we do not know whether the item is disliked (negative feedback) by the user or if he has not been exposed to this particular item (missing feedback). When data is binarized, this problem is known as one-class completion matrix~\citep{sindhwani_one-class_2010,davenport_1-bit_2014} or positive-unlabeled learning~\citep{hsieh_pu_2015}.

To select items which are going to be proposed to each user, traditional practice is to rank for each user $u$ the items he has not consumed yet (i.e., $\{ i | y_{ui} =0\}$), according to the score defined by:
\[\operatorname{score}_{ui}=[\hat{\W}\hat{\Hbf}^T]_{ui},\]
where $\hat{\W}$ are the estimated user preferences and $\hat{\Hbf}$ are the estimated item attributes. 

\subsection{Related Works}

\paragraph{Negative Binomial Regression.}
Regression for count data based on the Poisson distribution has been considered by \cite{gardner_regression_1995}. It has been augmented by a latent variable $a$ to model over-dispersion in \cite{lawless_negative_1987,hilbe_negative_2011,zhou_lognormal_2012}:
\begin{align}
y_i\sim\operatorname{Pois}(a_i\exp (\mathbf{x}_i^T\mathbf{b})),
\end{align}
where $y_i\in \mathbb{N}$ is the response variable, $\mathbf{x}_i$ is the covariate vector for sample $i$ and $\mathbf{b}$ is the vector of regression coefficients. When $a_{i}$ is given a gamma prior and marginalized, we get NB regression:
\begin{align}
y_i&\sim\operatorname{NB}\left(\alpha,\dfrac{1}{1+\alpha /\exp (\mathbf{x}_i^T\mathbf{\mathbf{b}})}\right). \label{eqn:nbreg}
\end{align}
Equation~\eqref{eqn:nbreg} defines a generalized linear model \citep{McCullagh1989} in which the data expectation is not linear in the parameters. We work instead with the mean-parametrized form of Equation~\eqref{NB}, which is more natural to the MF/dictionary learning setting. Furthermore, we also learn the ``covariates'' (similar to $\W$ in our case) and assume all variables to be nonnegative.

\paragraph{Outliers Modeling.}
As we explained, $\A$ can be interpreted as a variable that accounts for outliers. \citet{fevotte_nonlinear_2015} proposed a different way for handling outliers in NMF models and in particular in Poisson factorization (in the context of hyperspectral image unmixing). The outliers are modeled with an additive latent variable. The data is assumed Poisson-distributed with expectation $[\W\Hbf^T]_{ui} + s_{ui}$ where $s_{ui}$ is imposed to be sparse and non-negative. The non-negativity implies that only unexpectedly high data values can be captured with such a model. Here, we propose a multiplicative modeling of outliers which can explain unexpectedly high or low values.

\paragraph{Weighted MF.} Several methods have been proposed to lift the ambiguity associated to zero values in implicit feedbacks. A popular technique consists of considering all the zeros as negative feedbacks and down-weigh their importance during the inference. This is known as weighted MF (WMF)~\citep{hu_collaborative_2008,pan_one-class_2008}. In~\citet{liang_modeling_2015}, the authors show that WMF is a special case of models with exposure variables (like ours).

\paragraph{Exposure Modeling.}
NBMF can be cast as a particular instance of the following general model:
\begin{align}
\A&\sim p(\A;\Theta) \\ 
y_{ui}|a_{ui} &\sim \operatorname{Pois}(a_{ ui}[\W\Hbf^T]_{ui}),
\end{align}
where $p(\A;\Theta)$ is a distribution governed by its own parameters $\Theta$.

There are a few examples of such models in the literature, as described next. 
\begin{itemize}
  \item When $\A$ is deterministic with $\forall (u,i), a_{ui} = 1$, we recover the well-known PF model \citep{canny_gap:_2004,buntine_discrete_2006,cemgil_bayesian_2009,ma_probabilistic_2011,gopalan_scalable_2015-1}. 
  \item Zero-inflated models: in~\citet{simchowitz_zero-inflated_2013} $a_{ui}$ is drawn from a Bernoulli distribution: $a_{ui}\sim \mathcal{B}(\mu)$. Marginalizing out this latent variable leads to zero-inflated Poisson distribution~\citep{lambert_zero-inflated_1992}: 
  \[y_{ui} \sim (1-\mu)\delta_0 + \mu\operatorname{Pois}([\W\Hbf^T]_{ui}).\]
  In practice, it appears that the Bernoulli distribution puts too much weight on $0$. The gamma distribution offers a softer alternative. \citet{simchowitz_zero-inflated_2013} also proposes more sophisticated hierarchical models for $\mu$ (which becomes $\mu_{ui}$) to include external sources of knowledge (social network or geographical informations). Such ideas could also be incorporated in our setting.
  \item Coupled compound PF: \citet{basbug_coupled_2017} consider matrix completion with PF and missing-not-at-random phenomenas. Their approach relies on the following assumption
   \[a_{ui}\sim \operatorname{Pois}([\U\V^T]_{ui}).\]
 which is more restrictive (in terms of support and structure) than our proposal. The general purpose is also different. The general model is not conjugate anymore and sophisticated approximation are needed to infer the latent variables.
  \item Random graphs: in~\citet{paquet_one-class_2013}, the exposure is modeled with bipartite random graphs and it is arbitrarily assumed that half of the unconsumed items are missing feedbacks.
\end{itemize}

\section{Maximum Likelihood Estimation}
\label{sec:mle}

Before turning to more sophisticated Bayesian inference procedure, we study maximum likelihood estimation in the proposed model~\eqref{NB} and discuss the data fitting term that arises from our model.

\subsection{A New Divergence}

The maximum likelihood (ML) estimator of $\W$ and $\Hbf$ is obtained by minimizing the objective function defined by:
\begin{align}
C_{\text{ML}}(\W,\Hbf) &= - \log p(\Y;\W,\Hbf) \\
&= \sum_{ui} d_\alpha (y_{ui}|[\W\Hbf^T]_{ui}) + cst, \label{eqn:nbnmf}
\end{align}
where $cst$ is a constant with respect to (w.r.t.) $\W$ and $\Hbf$ and
\bal{d_\alpha (a|b) = a\log\left(\dfrac{a}{b}\right) - (\alpha+a)\log\left(\dfrac{\alpha + a}{\alpha + b}\right).}

\begin{figure}[t]
\centerline{\includegraphics[height=6cm]{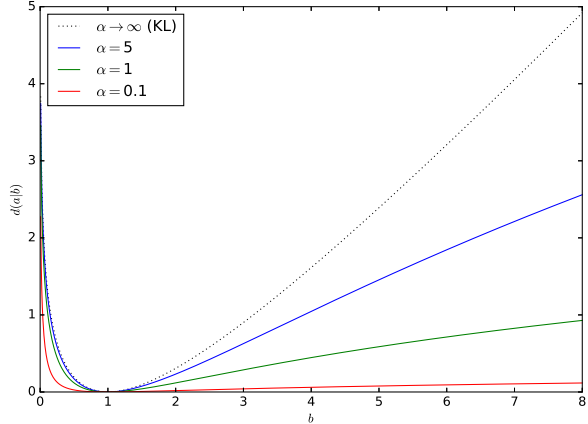}}
\caption{The ``mean-parametrized NB'' divergence $d_\alpha(a|b)$ for $b=1$ and four values of $\alpha$.}
\label{fig:div_nbf}
\end{figure}

$d_\alpha$ is the divergence associated to the mean-parametrized NB distribution with fixed dispersion coefficient $\alpha$. It is displayed in Figure~\ref{fig:div_nbf} for various values of $\alpha$. To the best of our knowledge, this divergence does not have a name nor  corresponds a well-known case from the literature. As expected, we recover in the limit case the generalized Kullback-Leibler divergence associated with the Poisson distribution: 
\[\lim\limits_{\alpha\to\infty}d_{\alpha}(a|b)=a\log\left(\dfrac{a}{b}\right) - a +b =d_{KL}(a|b).\]

\subsection{Block-Descent Majorization-Minimization}
As it turns out, maximum likelihood reduces to minimization of
\bal{
C(\W,\H) = D_{\alpha}(\ve{Y} | \ve{W} \ve{H}^{T}) \label{eqn:nbnmf2}
}
where $D_{\alpha}( \cdot | \cdot)$ is the entry-wise matrix divergence induced by $d_{\alpha}(\cdot|\cdot)$. Equation~\eqref{eqn:nbnmf2} defines a new NMF problem. A standard approach to minimize $C(\W,\H)$ is alternate block-descent optimization in which $\W$ and $\H$ are updated in turn until convergence to a stationary point (which may not be the global solution owing to the non-convexity of $C(\W,\H)$). The individual updates for $\W$ and $\H$ can be obtained using majorization-minimization (MM) like in many NMF cases, and such as NMF with the $\beta$-divergence \citep{fevotte_algorithms_2011}. The roles of $\W$ and $\H$ can be exchanged by transposition ($\Y \approx \W \H^{T} $ is equivalent to $\Y^{T} \approx \H \W^{T}$) and we may for example address the update of $\H$ given $\W$. MM amounts to optimizing an upper bound $G(\H|\bar{\H})$ of $C(\W,\H)$, constructed so as to be tight at the current iterate $\bar{\H}$ ($G(\bar{\H}|\bar{\H})=C(\W,\bar{\H})$). This produces a descent algorithm where the objective function is decreased at every iteration \citep{hunt04}.

In our setting, a tight upper bound can be constructed by majorizing the convex and concave parts of $C(\W,\H)$, following the approach proposed in  \cite{fevotte_algorithms_2011} for NMF with the $\beta$-divergence. The convex part (terms in $-\log(x)$) may be majorized using Jensen's inequality. The concave part (terms in $\log(x+c)$) can be majorized using the tangent inequality. This procedure leads to the following multiplicative update that preserves nonnegativity given positive initializations:
\bal{ h_{ik} = \bar{h}_{ik}\ \dfrac{\sum_u \dfrac{y_{ui}}{[\W\bar{\Hbf}^T]_{ui}} w_{uk}}{\sum_u\dfrac{y_{ui}+\alpha}{[\W\bar{\Hbf}^T]_{ui}+ \alpha} w_{uk}}. \label{eqn:nmfH} }
Similarly, the update for $\W$ is given by:
\bal{ w_{uk} = \bar{w}_{uk}\ \dfrac{\sum_i \dfrac{y_{ui}}{[\bar{\W}\Hbf^{T}]_{ui}} h_{ik}}{\sum_i\dfrac{y_{ui}+\alpha}{[\bar{\W}\Hbf^{T}]_{ui}+ \alpha} h_{ik}}. \label{eqn:nmfW}}
As expected, the multiplicative updates of KL-NMF are obtained in the limit ${\alpha \to \infty}$. An other way to obtain the updates described by Equations~\eqref{eqn:nmfH} and~\eqref{eqn:nmfW} is to use an Expectation-Minimization (EM) algorithm based the auxiliary variables $\ve{C}$ introduced in Section~\ref{bayesian_formulation}.

\section{Bayesian Negative Binomial Matrix Factorization} \label{sec:bayesian nbmf}

\subsection{Bayesian Formulation} \label{bayesian_formulation}
We describe a Bayesian formulation of NBMF based on gamma priors on both $\W$ and $\Hbf$. The gamma distribution is a natural candidate because it is conjugate with the Poisson distribution. Moreover, adding a gamma prior with a shape parameter lower than $1$ is known to induce sparsity, which is an hypothesis often used in CF~\citep{ma_probabilistic_2011}. As such, we set:
\begin{align}
w_{uk}&\sim\mathcal{G}(\alpha^W,\beta^W), \\
h_{ik}&\sim\mathcal{G}(\alpha^H,\beta^H). 
\end{align}
It is useful in PF to introduce latent components $c_{uik}$ that derive from the superposition property of Poisson variables~\citep{cemgil_bayesian_2009}. It leads to the composite model described by:
\begin{align}
a_{ui}&\sim \mathcal{G}(\alpha,{\alpha}), \\ 
c_{uik}|\A,\W,\Hbf &\sim \operatorname{Pois}(a_{ui}w_{uk}h_{ik}),\\
y_{ui}&=\sum_k c_{uik}.
\end{align}
We denote by $\cbf_{ui}$ the vector $[c_{ui1},...,c_{uiK}]^T$ and by $\C$ the tensor of size $U\times I\times K$ with coefficients $c_{uik}$. We denote by $z=\{ \C,\A,\W,\Hbf \}$ the set of latent variables, $\theta^{\text{shape}} = \{ \alpha, \alpha^W, \alpha^H \}$ and $\theta^{\text{rate}} = \{\beta^W,\beta^H\}$ the set of shape and rate hyperparameters, respectively. We denote by $\theta = \theta^{\text{shape}}\cup \theta^{\text{rate}}$ the set of all hyperparameters.

\begin{figure}[t]
\centering
\tikzset{
  font={\fontsize{8pt}{12}\selectfont}}
\begin{tikzpicture}
\tikzstyle{main}=[circle, minimum size = 6.5mm, thick, draw =black!80, node distance = 6.5mm]
\tikzstyle{connect}=[-latex, thick]
\tikzstyle{box}=[rectangle, draw=black!100]

\node[main, fill = white!100] (w) [label=center:$w_{uk}$] { };
\node[main] (c) [right=of w,label=center:$c_{uik}$] { };
\node[main, fill = black!30] (x) [below=of c,label=center:$y_{ui}$] { };
\node[main] (h) [right=of c,label=center:$h_{ik}$] { };
\node[main] (v) [above=of c,label=center:$a_{ui}$] {};

\path (w) edge [connect] (c)
    (h) edge [connect] (c)
    (v) edge [connect] (c)
    (c) edge [connect] (x);

\node[rectangle, inner sep=4mm,draw=black!100, fit= (w) (h) (c), xshift=1.5mm] {};
\draw (-0.4,-0.5) node{$K$};

\node[rectangle, inner sep=4mm,draw=black!100, fit= (w) (v) (x) (c), yshift=0.75mm] {};
\draw (-0.5,-1.8) node{$U$};

\node[rectangle, inner sep=4mm,draw=black!100, fit= (h) (v) (x) (c), yshift=-0.75mm] {};
\draw (3.2,-1.9) node{$I$};
\end{tikzpicture}

\caption{Graphical representation of Bayesian NBMF with auxiliary latent variables. Hyperparameters are not represented here.}
\end{figure}
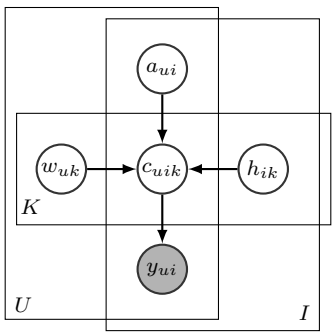

\subsection{Variational Inference}

\subsubsection{Mean-Field Approximation}
Bayesian inference revolves around the characterization of the posterior distribution $p(z|\Y;\theta)$.
Unfortunately, this posterior is intractable in our model. We propose to compute an approximation based on variational inference (VI) \citep{jordan_introduction_1999}. The main idea behind VI is to approximate the posterior by a simpler distribution $q$. The inference problem becomes an optimization problem described by:
\[q = \argmin_{q\in\mathcal{F}} \operatorname{KL}(q| p_{z|\Y}),\]
where KL here refers to the KL divergence between distributions and where $\mathcal{F}$ is a family of distributions. The log-likelihood can be decomposed using Jensen inequality as:
\[ \log p(\Y;\theta) = \operatorname{ELBO}(q,\theta) + \operatorname{KL}(q| p_{z|\Y}), \]
where
$\operatorname{KL}$ is here the KL divergence between distributions defined by $\operatorname{KL}(q| p_{z|\Y}) =\int_z q(z) \log \frac{q(z)}{p(z|\Y;\theta)}dz$,
and ELBO is the expected lower bound defined by: $\operatorname{ELBO}(q,\theta) = \int_z q(z) \log \frac{p(\Y,z;\theta)}{q(z)}dz$. It is a lower bound of the log likelihood since $\operatorname{KL}(q| p_{z|\Y})\geq 0$. Minimizing the KL divergence is therefore equivalent to maximize the ELBO, which is simpler in practice.

A common choice for $\mathcal{F}$ is the mean-field family which assumes $q$ to be fully factorizable:
\[q(z)= \prod_{ui}q(\cbf_{ui})q(a_{ui})\prod_{uk}q(w_{uk})\prod_{ik}q(h_{ik}).\]

\subsubsection{Coordinate Ascent for Variational Inference}
We use a coordinate ascent for VI (CAVI) algorithm to maximize the ELBO. The CAVI algorithm consists of sequentially optimizing each of the variational parameters keeping the others fixed. It appears that using the mean field family and CAVI algorithm leads to the closed-form solution~\citep{bishop_pattern_2006}:
\begin{align}
\log q(x) &= \langle \log p(x|z_{-x})\rangle_{q(z_{-x})} + cst,
\end{align}
where $\langle .\rangle_q$ denotes expectation under distribution $q$ and $z_{-x}$ denotes the set of latent variables $z$ excluding the variable $x$.

Our model is conjugate and we can write the following conditional posteriors:\footnote{Note that a Gibbs algorithm could also be developed based on these conditional posteriors.}
\begin{align*}
&\cbf_{ui}|z_{-\cbf_{ui}}\sim \operatorname{Mult}(y_{ui},\dfrac{w_{uk}h_{ik}}{[\W\Hbf^T]_{ui}}), \\
&a_{ui}|z_{-a_{ui}}\sim \mathcal{G}(\alpha+y_{ui},~\alpha + [\W\Hbf^T]_{ui}),\\
&w_{uk}|z_{-w_{uk}}\sim\mathcal{G}(\alpha^W+\sum_i c_{uik},~\beta^W + \sum_i a_{ui}h_{ik}), \\
&h_{ik}|z_{-h_{ik}}\sim\mathcal{G}(\alpha^H+\sum_u c_{uik},~\beta^H + \sum_u a_{ui}w_{uk}).
\end{align*}

This leads to the following closed-form expressions for the variational distribution: 
$q(\cbf_{ui})= \operatorname{Mult}(y_{ui},\mathbf{\phi}_{ui})$, 
$q(a_{ui}) = \mathcal{G}(\tilde{\alpha}^A_{ui},~\tilde{\beta}^A_{ui})$,
$q(w_{uk}) = \mathcal{G}(\tilde{\alpha}^W_{uk},~\tilde{\beta}^W_{uk})$ and
$q(h_{ik})= \mathcal{G}(\tilde{\alpha}^H_{ik},~\tilde{\beta}^H_{ik})$, where
\smallskip
\begin{align}
&\phi_{uik}\propto \exp\left(\langle \log w_{uk}\rangle_q + \langle \log h_{ik}\rangle_q\right), \label{eq:update_c}\\
&\tilde{\alpha}_{ui} = \alpha+ y_{ui},~
\tilde{\beta}_{ui} = \alpha+ \sum_k \langle w_{uk}\rangle_q\langle h_{ik}\rangle_q, \label{eq:update_a}\\
&\tilde{\alpha}^W_{uk} = \alpha^W+\sum_i y_{ui} \phi_{uik},~
\tilde{\beta}^W_{uk} = \beta^W+ \sum_i \langle a_{ui}\rangle_q\langle h_{ik}\rangle_q, \label{eq:update_w}\\
&\tilde{\alpha}^H_{ik} = \alpha^H+\sum_u y_{ui} \phi_{uik},~
\tilde{\beta}^H_{ik} = \beta^H + \sum_u \langle a_{ui}\rangle_q\langle w_{uk}\rangle_q. \label{eq:update_h}
\end{align}
When $q(x) = \mathcal{G}(\alpha,\beta)$ with have 
$\langle x\rangle_q=\frac{\alpha}{\beta}$ and 
$\langle \log x\rangle_q = \Psi(\alpha)- \log(\beta)$, where $\Psi$ is the digamma function defined by: $\Psi(x)=\frac{d\log\Gamma}{dx}(x)$.

Note that we only need to infer $\cbf_{ui}$ on pairs $(u,i)$ such that $y_{ui}>0$.

\subsubsection{Hyperparameter Estimation}

Let $\lambda\in \mathbb{R}_+$, $\tilde{\theta}=\theta^{\text{shape}}\cup\{\lambda\beta^W,\lambda^{-1}\beta^H\}$ and $\tilde{q}$ be such that $\tilde{q}(\W)=\prod_{uk}\mathcal{G}(\tilde{\alpha}^W_{uk},\lambda\tilde{\beta}^W_{uk})$ and $\tilde{q}(\Hbf)=\prod_{ik}\mathcal{G}(\tilde{\alpha}^H_{ik},\lambda^{-1}\tilde{\beta}^H_{ik})$. It can be easily checked that both the likelihood and the ELBO are scale invariant:
\begin{align}
p(\Y;\tilde{\theta}) &= p(\Y;\theta), \\
\operatorname{ELBO}(\tilde{q},\tilde{\theta}) &= \operatorname{ELBO}(q,\theta). 
\end{align}
Consequently, we can fix $\beta^W$ and maximize the lower bound $\operatorname{ELBO}(q,\theta)$ w.r.t. $\beta^H$. We set $\beta^W=\alpha^W$ such that $\mathbb{E}[w_{uk}]=1$. Moreover, we have $\mathbb{E}[a_{ui}]=1$ and the scale information is therefore only carried by $\beta^H$.  Optimizing $\beta^{H}$ leads to:
\begin{equation}\label{eq:update_hyper}
\beta^H=\dfrac{\alpha^H}{\sum_{ik}\langle h_{ik} \rangle_q / UI}.
\end{equation}

The shape hyperparameters $\theta^{\text{shape}}=\{\alpha, \alpha^W, \alpha^H\}$ could also be updated with a Newton-Raphson method \citep{cemgil_bayesian_2009} but we treat them here like user-defined constant. 
\subsection{Algorithm}
The complete CAVI algorithm is described in Algorithm~\ref{algo:VI}. It is initialized with random values and stopped once the relative increment of the ELBO gets lower than a chosen parameter $\tau$.

\begin{algorithm}[tb]
\SetKwData{Left}{left}\SetKwData{This}{this}\SetKwData{Up}{up}
\SetKwFunction{Union}{Union}\SetKwFunction{FindCompress}{FindCompress}
\SetKwInOut{Input}{Input}\SetKwInOut{Output}{Output}
\Input{$\Y$, $K$, $\theta$}
\Output{$q(\C,\A,\W,\Hbf)$}
\BlankLine
Initialize expectations: $\langle w_{uk}\rangle_q$, $\langle h_{ik}\rangle_q$, $\langle \log w_{uk}\rangle_q$ and $\langle \log h_{ik}\rangle_q$\\
\Repeat{ELBO converges}
{
	for each pair $(u,i)$ such that $y_{ui}>0$: Eq.~\eqref{eq:update_c}\\
	for each pair $(u,i)$: Eq.~\eqref{eq:update_a}\\
	for each user $u$: Eq.~\eqref{eq:update_w}\\
	for each item $i$: Eq.~\eqref{eq:update_h}\\
	optimize hyperparameters: Eq.~\eqref{eq:update_hyper}
}
 \caption{CAVI algorithm}
 \label{algo:VI}
\end{algorithm}

\subsection{Expected Predictive Posterior} \label{sec:score}
We use the expected predictive posterior to assess what users will like in the future. The predictive posterior is given by the probability $p(y^*_{ui}|\Y)$  of a new observation $y^*_{ui}$ given the observed data $\Y$. Its expected value can be approximated using the variational distribution $q$:
\begin{align}
\mathbb{E}[y^*_{ui}|\Y] &= \sum_k \langle w_{uk}\rangle_q\langle h_{ik}\rangle_q.
\end{align}
Note that we recover the score introduced empirically in Section~\ref{sec:reco task}.

\section{Application to Recommender Systems}

\subsection{Experimental Setup}  \label{sec:exp setup}

\paragraph{Dataset.} 
We consider the Taste Profile dataset, provided by The Echo Nest~\citep{bertin-mahieux_million_2011}. This dataset contains the listening history of users in the form of song play counts. 

\begin{figure}[t]
\centerline{\includegraphics[height=3.5cm]{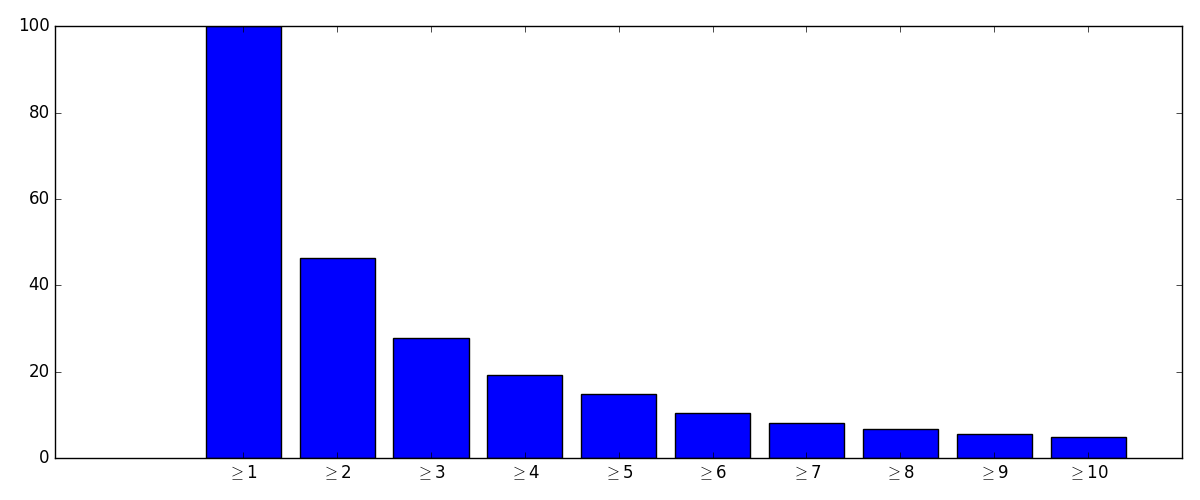}}
\caption{Cumulative histogram of non-zero values of the Taste Profile dataset. Listening count greater or equal to $2$ ($10$) represent $46\%$ ($5\%$ respectively) of non-zero values. }
\label{fig:histo_cumul}
\end{figure}

As in \citet{liang_modeling_2015}, we select a subset of the original data by only keeping users who listened to at least 20 different songs, and songs which have been listened to at least by 50 different users. This leads to a dataset with a number of users $U=1509$ and a number of items $I=805$. We summarize these play counts in a matrix $\Y\in\mathbb{N}^{U\times I}$. The percentage of non-zero values in our subset is $5\%$. The cumulative histogram of non-zero values of $\Y$ is presented in Figure~\ref{fig:histo_cumul}. We can see that about half of the non-zero listening counts are ones ($54\%$). We have to be careful with such counts. As explained in~\citet{hu_collaborative_2008}, the value of implicit feedbacks indicates confidence and not preference. In fact, the low values are more sensitive to noise (an item listened to only once can be disliked).

To evaluate our algorithm, we randomly divide our observed matrix $\Y$ into two matrices $\Y^{\text{train}}$ and $\Y^{\text{test}}$. $80\%$ of the non-zero values of $\Y$ correspond to those in $\Y^{\text{train}}$, while the other $20\%$ to those in $\Y^{\text{test}}$. The other values are set to zero to preserve the ambiguity between negative and missing feedbacks. We infer our model with $\Y^{\text{train}}$ using the algorithm described in~Alg.~\ref{algo:VI}. We propose to each user a personalized list of recommendation. This list is constructed by decreasing order of the score defined in Section~\ref{sec:score}, with items already consumed placed at the end.

\paragraph{Evaluation Metric.}

We use the normalized discounted cumulative gain (NDCG) to evaluate and compare the performance of the different models. NDCG is a metric often used in information retrieval to evaluate ranking lists of predictions. 

For each user, we calculate the discounted cumulative gain (DCG), defined by:
\[\operatorname{DCG}_u = \sum_i \dfrac{2^{\operatorname{rel}(u,i)}-1}{\log_2(\operatorname{rank}_u(i)+1)}, \]
where $\operatorname{rel}(u,i)$ is the ground-truth relevance of item $i$ for user $u$, and $\operatorname{rank}_u(i)$ is the rank of item $i$ in the recommendation list of user $u$. For example, for a user $u$, if the first item predicted in the list is the item $i'$, then we have $\operatorname{rank}_u(i')=1$. The denominator penalizes relevant items which are at the end of the proposed list. It accounts for the fact that a user will only browse the beginning of the list, and will not pay attention to items which are ranked at the end.

We propose two different choices for the ground-truth relevance:
\begin{itemize}
	\item $\operatorname{rel}_A(u,i)=y^{\text{test}}_{ui}$. More weight is given to items which have been listened to a high number of times. This choice respects the fact that low listening counts reflect a preference with low confidence.
	\item $\operatorname{rel}_B(u,i)= \mathds{1}[y^{\text{test}}_{ui}\geq s]$, where $s$ is a fixed threshold. When $s=0$, we recover the classic NDCG metric for binary data. When $s>0$, we focus only on items which have been listened to at least $s$ times. It totally ignores listening counts lower than $s$ for which the confidence may not be high enough. 
\end{itemize}

DCG does not have a fixed scale making it hard to analyze. We can normalize it with: 
\[ \operatorname{NDCG}_u = \frac{\operatorname{DCG}_u}{\operatorname{IDCG}_u}, \] 
where $\operatorname{IDCG}_u$ is the ideal DCG. It corresponds to the DCG score of an oracle which ranks perfectly its recommendation list (by decreasing order of $y^{\text{test}}_{ui}$). We report in the next section the average NDCG over all users.

\paragraph{Compared Methods.}
We compare NBMF with two versions of PF \citep{gopalan_scalable_2015-1}. One with pre-processing stage where we binarize the data $\Y^{train}$, and one without. For all models, we set $\alpha^W = \alpha^H = 1$, $\beta^W = \alpha^W$ and we learn $\beta^H$. Moreover, for NBMF, we set $\alpha = 1$. We fix the converge rate to $\tau = 10^{-5}$. All the algorithms are run 5 times with random initializations.

\subsection{Results}

\paragraph{Prediction Results.}

\begin{figure}[t]
\centerline{\includegraphics[height=5.5cm]{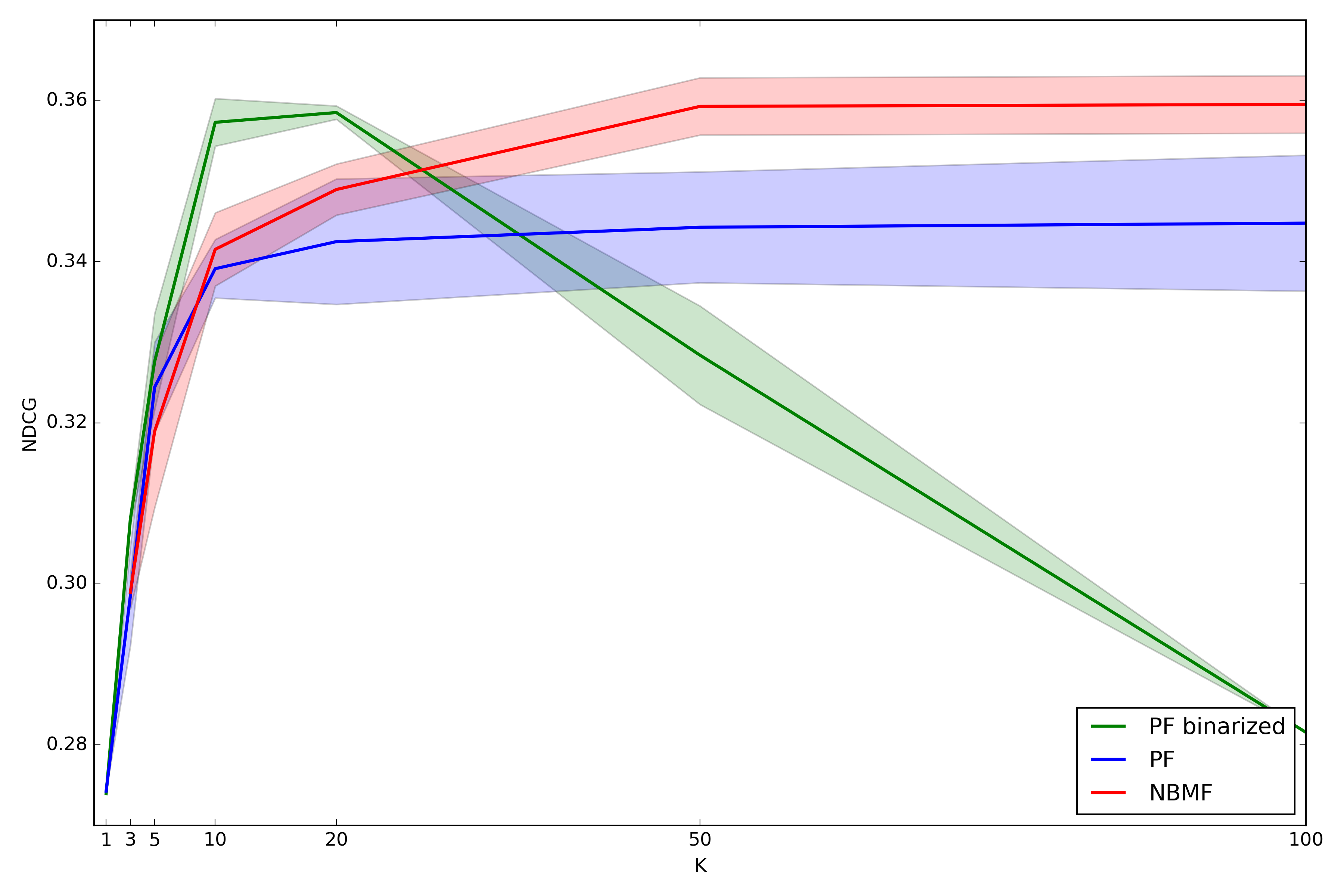}}
\caption{NDCG with ground truth relevance $\operatorname{rel}_A(u,i)=y^{\text{test}}_{ui}$ w.r.t. the number of latent factors $K$. }
\label{fig:map_k}
\end{figure}

\begin{figure}[t]
\centerline{\includegraphics[height=5.5cm]{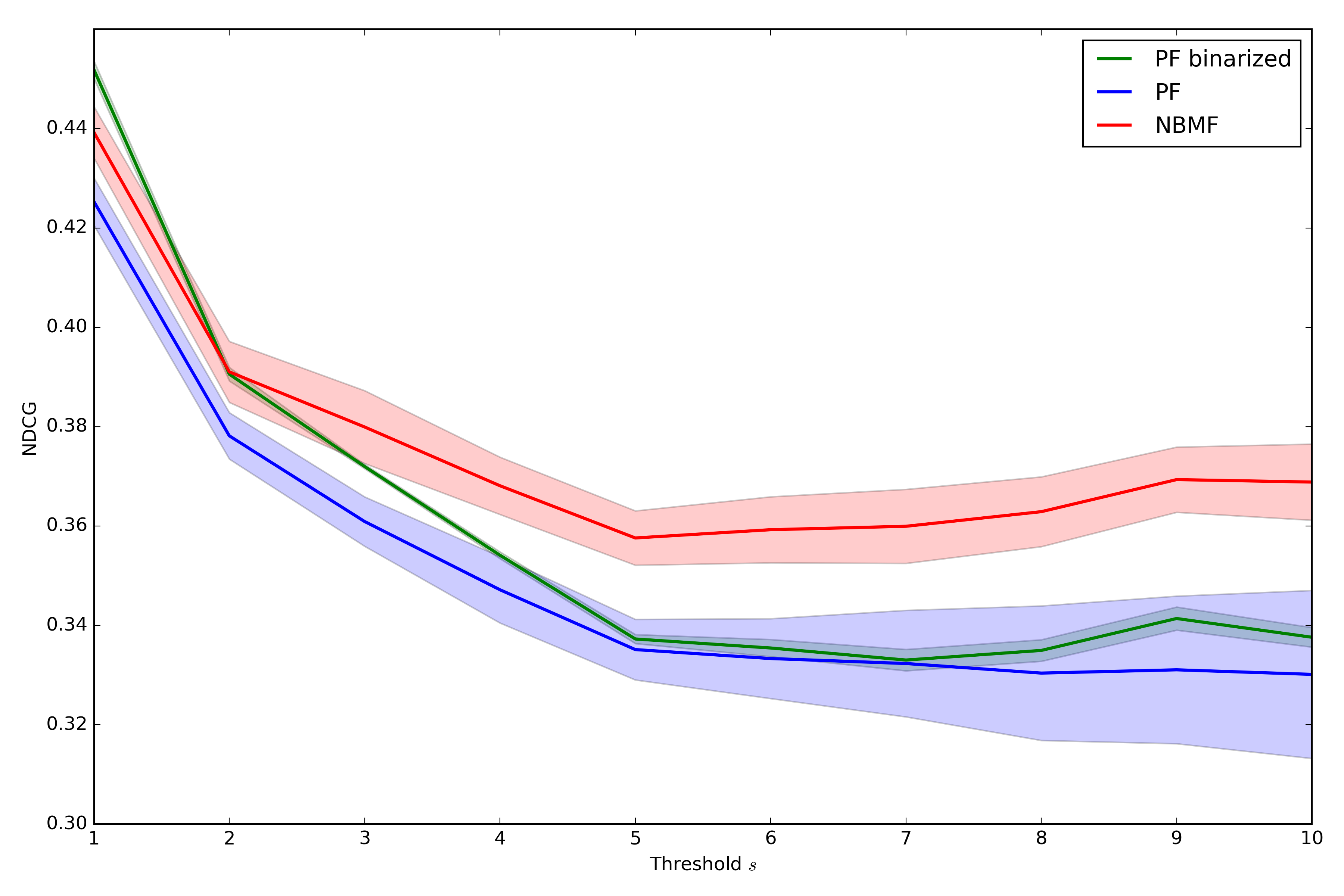}}
\caption{NDCG with ground truth relevance $\operatorname{rel}_B(u,i)= \mathds{1}[y^{\text{test}}_{ui}\geq s]$ w.r.t. the thresholding $s$.}
\label{fig:map threshold}
\end{figure}

Figure~\ref{fig:map_k} displays the performances of each model according to the NDCG metric defined with $\operatorname{rel}_A$ w.r.t the number of latent factor $K$. We can see that NBMF clearly outperforms PF without pre-processing stage, and seems slightly better than PF on binarized data. The maximum score is achieved at $K=20$ for binarized PF, and $K=50$ for NBMF and PF. This makes sense that binarized PF needs less latent factors, because it only models the support and not the values of $\Y^{\text{train}}$. 

Figure~\ref{fig:map threshold} displays the performances of each model according to the NDCG metric defined with $\operatorname{rel}_B$ w.r.t the threshold $s$. We chose $K$ as previously described, i.e. $K=20$ for binarized PF, and $K=50$ for NBMF and PF. For $s=1$, binarized PF seems better since it only models the support of $\Y^{\text{train}}$. As explained in Section~\ref{sec:exp setup}, low values of $\Y^{\text{train}}$ are very sensitive to noise. Thresholding to $s>1$ allows for more robustness in the NDCG metric. For such thresholding, we see that NBMF presents the best performances. 

\paragraph{Exploratory Analysis.} 

\begin{figure}[h]
\centerline{\includegraphics[height=6.4cm]{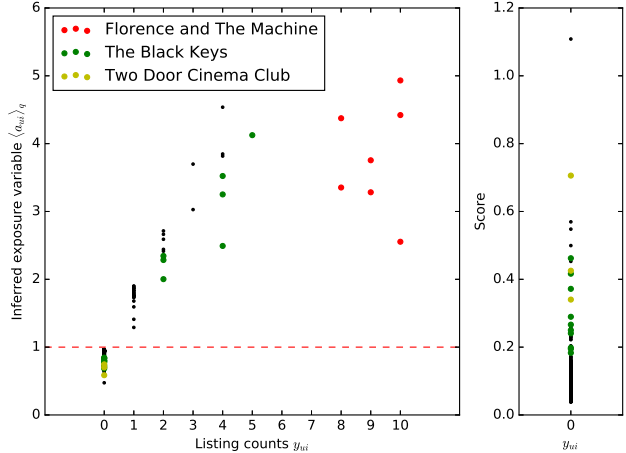}}
\caption{Example of the inferred exposures and recommendation list for a particular user. On the left, the plot illustrates the inferred exposures w.r.t. the listening counts for a particular user. The points represent all the songs present in the train set $\Y^{\text{train}}$, the x-axis is the number of play counts ($y_{ui}$) and the y-axis is the inferred expected value of the exposure variable ($\langle a_{ui} \rangle_q$). The red dashed line $\langle a_{ui} \rangle_q = 1$ corresponds to standard consumption behavior. Each item above this line is ``over-consumed'', and each item under this line is ``under-consumed'' (see Section~\ref{sec:interpretation A}). On the right, the plot represents the score of the songs not listened by the user yet.}
\label{fig:explore A}
\end{figure}

Figure~\ref{fig:explore A} illustrates the listening habits of a particular user. The user has mainly listened to two bands: Florence and the Machine (red points) and The Black Keys (green points). Both are two recent ``indie rock'' bands from the 2000s. Both are considered over-consumed and $a_{ui}$ therefore down-weights their influence. Note that $a_{ui}$ seems to be similar for both bands but, as the user has more listened to Florence and the Machine, this band will keep a stronger influence on his inferred preferences.

We now look at the unconsumed items of this user ($y_{ui}=0$). We see that the items are discriminated by their exposure values. The user seems to like the band The Black Keys, but there are still some songs from this band he has not listened to yet. Exposure for these items are low, meaning that he has not been exposed to it (the songs can be from an album he does not know for example) rather than  he does not like them.

The recommended items correspond to items with high score (see Section~\ref{sec:score}). The first recommended item (point with the higher score) is ``I Wanna Be Sedated'' from The Ramones. This is an American punk rock band from the 1970s, which inspired a lot of current music bands (such as The Black Keys). It seems to be coherent to recommend such an item the user certainly does not know but that he could like. Similarly, an other ``indie rock'' band that the user does not seem to know, Two Door Cinema Club, is recommended to the user.

\section{Conclusion}
In this paper, we introduced a new matrix factorization technique for over-dispersed data, NBMF. NBMF is an extension of PF where a latent additional variable models the exposure. It leads to finer recommendations on the Taste Profile dataset.
Future work will consist in proposing a faster algorithm based on stochastic VI~\citep{hoffman_stochastic_2013}. Another exciting perspective would be to add temporal information to the exposure variable. In particular, a Markov chain structure could be exploited~\citep{cemgil_conjugate_2007,fevotte_non-negative_2013,jerfel_dynamic_2016}. Adding structure to the latent exposure could also improve the recommendation~\citep{basbug_hierarchical_2016}. Last but not least, NBMF could to be applied to a wider range of data and applications that involve integer-valued, such as bags of words or images.

\newpage
\subsubsection*{References}

\begingroup
\renewcommand{\section}[2]{}

\endgroup

\end{document}